\pdfoutput=1
\documentclass[10pt,twocolumn,letterpaper]{article}
\usepackage[pagenumbers]{cvpr}
\usepackage{microtype}
\definecolor{cvprblue}{rgb}{0.21,0.49,0.74}
\usepackage[pagebackref,breaklinks,colorlinks,allcolors=cvprblue]{hyperref}
\def\confName{CVPR}
\def\confYear{2027}

\title{CUFormer: Modality-Aware CT-Ultrasound Point Cloud Registration for Orthopedic Navigation}
\author{Lixing Tan$^{1}$\quad Hongye Hao$^{1}$\quad Tianrun Gu$^{1}$\quad Qing Xia$^{1,*}$\quad Yuting Guo$^{2}$\quad Shuai Li$^{1}$\quad Aimin Hao$^{1}$\\[0.4em]
$^{1}$Beihang University\qquad $^{2}$Beijing Information Science and Technology University\\[0.3em]
{\normalsize Project page: \url{https://lextran.github.io/CUFormer/}}\\[0.2em]
{\small $^{*}$Corresponding author: \href{mailto:xiaqing@buaa.edu.cn}{\texttt{xiaqing@buaa.edu.cn}}}}
\begin{document}

\twocolumn[{%
\renewcommand\twocolumn[1][]{#1}
\maketitle
\begin{center}
    \centering
    \captionsetup{type=figure}
    \includegraphics[width=\textwidth]{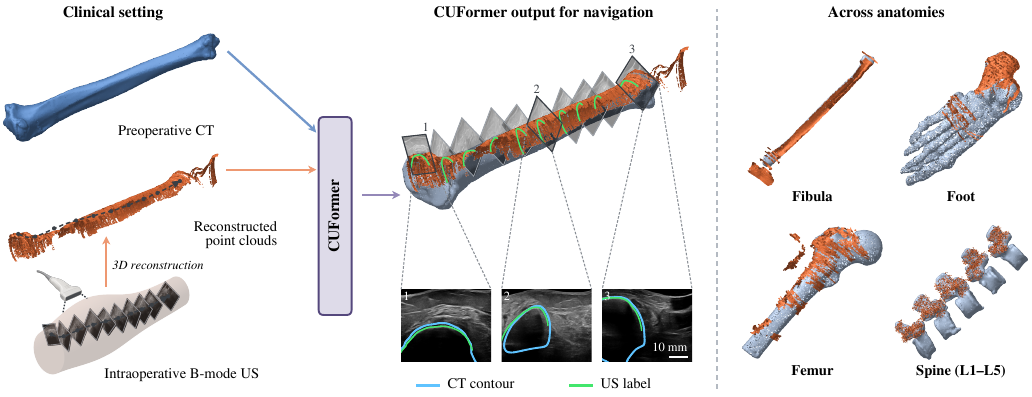}
    \captionof{figure}{CUFormer registers intraoperative US bone surfaces to preoperative CT for radiation-free orthopedic navigation. Tracked B-mode images are reconstructed into a partial point cloud and aligned with the CT model. CT-US results span the tibia, fibula, foot, and femur. The spine example adds CT-RGB-D registration. Both US and RGB-D point clouds provide partial bone-surface observations.}
    \label{fig:teaser}
\end{center}
}]

\begin{abstract}
Registering intraoperative Ultrasound (US) bone-surface point clouds to preoperative Computed Tomography (CT) models can support radiation-free orthopedic navigation. Differences in imaging physics produce distinct surface sampling patterns, complicating cross-modal feature matching. Bone strongly reflects ultrasound and casts an acoustic shadow, so US captures only the probe-facing bone surface and leaves most CT points unmatched. This partial coverage makes correspondences ambiguous on near-symmetric bones. We propose CUFormer, a modality-aware registration framework. We introduce a Modality-Aware Transformer (MAT) that encodes input modality through identity embeddings and updates features through modality-specific adapters within a shared geometric matcher. We model imaging visibility and use predicted reference-side visibility to weight coarse and fine correspondences. CUFormer achieves the lowest mean registration errors among evaluated methods on two CT-US datasets, UltraBones100k and UltraBonesHip, and the CT-RGB-D dataset SpineDepth. On UltraBones100k, it reaches mean rotation error 4.53$^\circ$, translation error 3.19\,mm, and Root Mean Squared Error (RMSE) 2.54\,mm, reducing RMSE by 65.8\% relative to the best-performing baseline by this metric. These results demonstrate CUFormer's effectiveness in aligning partial bone surfaces with preoperative CT across anatomical sites.
\end{abstract}

\section{Introduction}

Image-guided orthopedic surgery requires millimeter-level localization of bony anatomy. Preoperative Computed Tomography (CT) provides a complete, high-resolution bone model. Using this model for intraoperative guidance requires aligning it with the patient's current anatomy. Repeated intraoperative CT or fluoroscopy exposes patients and clinicians to ionizing radiation, motivating radiation-free intraoperative imaging. Ultrasound (US) provides real-time, non-invasive observations of local bone surfaces, which can be reconstructed as point clouds from tracked acquisitions. Registering these surfaces to the preoperative CT model establishes the spatial correspondence between the intraoperative anatomy and the preoperative reference for orthopedic navigation (Figure~\ref{fig:teaser}).

\begin{figure}[t]
    \centering
    \includegraphics[width=\columnwidth]{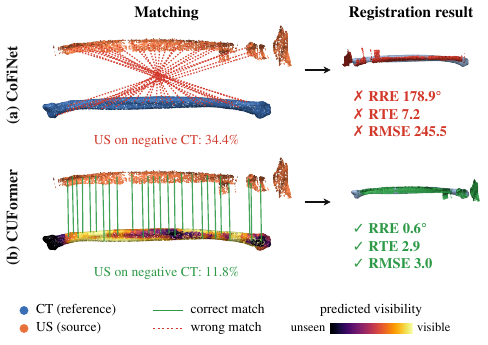}
    \caption{Comparison of a previous method and CUFormer. CT-US correspondence ambiguity on a fibula case. US is orange. Lines connect sampled US points to estimated CT positions: green indicates errors within 5\,mm of the Ground-Truth (GT) counterpart, and red indicates larger errors. (a) CoFiNet produces an end-for-end flip. (b) CUFormer uses MAT and visibility-guided matching, with CT colored by predicted visibility. Nearest-CT assignment places 34.4\% of all US points in negative regions for CoFiNet, 11.8\% for CUFormer, and 11.7\% for GT.}
    \label{fig:motivation}
\end{figure}

Representing both modalities as bone-surface point clouds removes image intensity differences~\citep{wu2025neuralbonereg}, but imaging physics still shapes the observed geometry. CT-US registration therefore faces two major challenges: modality-dependent surface sampling and highly asymmetric coverage. Modality-dependent sampling changes the local neighborhoods used to construct descriptors, making corresponding CT and US surface points harder to match. Coverage asymmetry also originates in imaging physics. Bone strongly reflects ultrasound waves and casts an acoustic shadow, so a typical tibial sweep captures only a thin, probe-facing surface strip and leaves most of the CT reference unobserved. This partial coverage makes correspondence estimation ambiguous along near-symmetric bone structures. Figure~\ref{fig:motivation}(a) shows a matcher placing the US source at the wrong end of the CT reference, producing an end-for-end flip.

Existing strategies can hardly address these challenges. Local refinement such as ICP~\citep{besl1992icp} requires a close initialization, and RANSAC-based estimation~\citep{fischler1981ransac} relies on putative correspondences, whose inlier ratio drops under partial observations. NeuralBoneReg (NBReg)~\citep{wu2025neuralbonereg} optimizes the transformation per case at test time, which limits real-time use, and sparse or limited surface observations may reduce its performance.

By contrast, learning-based matchers learn features across cases and register a new pair without per-case optimization, performing well on same-modality RGB-D benchmarks~\citep{qin2022geotransformer,yu2023rotation}. These methods treat the two inputs as scans from the same sensor. GeoTransformer~\citep{qin2022geotransformer} encodes both clouds with shared weights, leaving no mechanism to account for modality-dependent sampling. Low-overlap methods~\citep{huang2021predator,yu2021cofinet} model overlap as a symmetric relation between two partial scans, where non-overlap stems from viewpoint and scan range. In CT-US registration, the CT reference is complete, whereas US observes only part of the reference surface owing to the acoustic properties of bone. A CT-US matcher therefore needs features that distinguish the two modalities and correspondence selection that reflects reference-side coverage.

To address these challenges, we propose CUFormer. Its Modality-Aware Transformer (MAT) preserves a shared geometric matcher while providing explicit input identities and modality-specific feature updates. Learnable embeddings encode CT and US stream identities at coarse and fine resolutions. Gated bottleneck adapters provide modality-specific residual updates after shared attention blocks. A visibility module models imaging visibility through geometric overlap, predicting which CT regions are observed by US under overlap-derived supervision. The predicted visibility scores weight coarse and fine correspondence scores, guiding correspondence selection toward CT regions supported by the partial US observation. Figure~\ref{fig:motivation}(b) shows the result.

Our contributions are summarized as follows:
\begin{itemize}
    \item We propose \textbf{CUFormer}, a modality-aware framework for CT-US bone-surface registration in orthopedic navigation. It addresses modality-dependent surface sampling and asymmetric coverage within a single learned matcher that requires no per-case optimization.
    \item We introduce \textbf{MAT}, which integrates modality identity encoding and modality-specific residual adaptation into a shared geometric matcher.
    \item We develop \textbf{reference-side visibility-guided matching}, which complements MAT by modeling US imaging visibility on CT through geometric overlap and using the predicted visibility to weight coarse and fine correspondence scores.
    \item We evaluate CUFormer against state-of-the-art methods on (i) the public CT-US datasets UltraBones100k and UltraBonesHip, covering five structures: fibula, tibia, foot, femur, and pelvis; and (ii) the public CT-RGB-D dataset SpineDepth, covering multiple vertebrae. Ablations further assess the contribution of each component.
\end{itemize}

\section{Related Work}

\subsection{US-based Bone Imaging and Registration}

US is well suited to computer-assisted orthopedic surgery because it provides real-time imaging without ionizing radiation. However, extracting and registering bone surfaces from US remains challenging because of limited coverage and modality-specific imaging artifacts. For surface extraction, UltraBones100k~\citep{wu2025ultrabones100k} provides an automated labeling pipeline and a large-scale, CT-labeled US dataset, and UltraBoneUDF~\citep{wu2025ultraboneudf} reconstructs open bone surfaces from US using a neural unsigned distance function. For registration, surface-based pipelines commonly refine an initial alignment with ICP~\citep{besl1992icp}. NBReg~\citep{wu2025neuralbonereg} aligns multimodal bone surfaces by fitting a separate neural distance field to each patient's preoperative model and optimizing the transformation at test time. In contrast to such per-case optimization, CUFormer learns a single-pass matcher for CT-US registration with modality-aware features and complements its correspondence scores with predicted CT visibility.

\subsection{Learning-based Point Cloud Registration}
\begin{figure*}[htb]
    \centering
    \includegraphics[width=\textwidth]{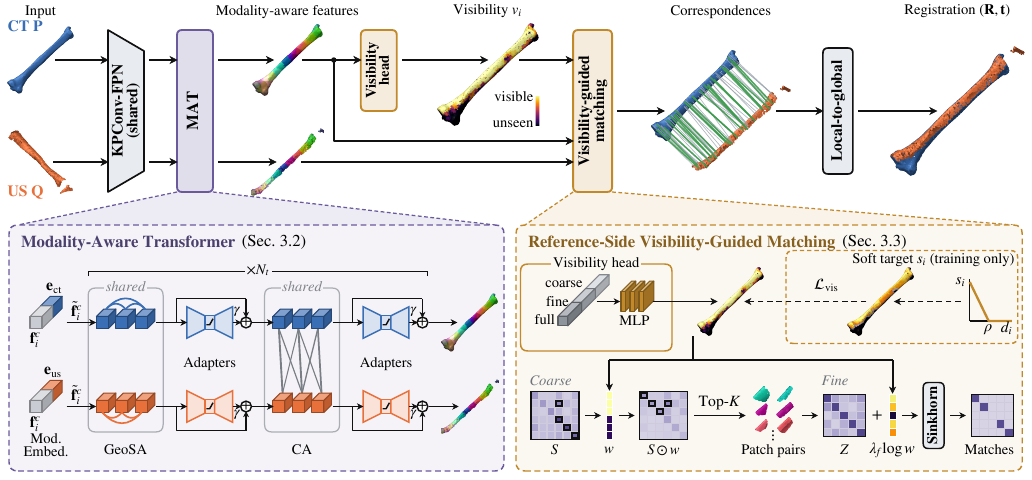}
    \caption{Overview of CUFormer. Top: a shared encoder and MAT produce modality-aware features. Reference-side visibility guides coarse and fine matching before rigid transformation estimation. Bottom left: modality embeddings are added to coarse features before attention and to fine features before matching. Separate CT/US gated adapters follow each shared self- and cross-attention block. Adapter activation icons denote ReLU. Bottom right: fused multi-resolution features predict CT visibility, supervised by overlap-derived soft targets. Visibility weights coarse scores before top-$K$ selection and biases fine scores before Sinkhorn.}
    \label{fig:overview}
\end{figure*}

The growing availability of large-scale datasets has advanced learning-based registration. These methods combine geometric features, coarse-to-fine correspondences, and robust transformation estimation~\citep{qi2017pointnet,choy2019fcgf,bai2020d3feat,wang2019dcp,yew2020rpmnet,choy2020dgr,bai2021pointdsc,yew2022regtr}. For limited overlap, Predator~\citep{huang2021predator} predicts overlap and matchability but does not consider the asymmetry between complete and partial coverage. CoFiNet~\citep{yu2021cofinet} builds keypoint-free hierarchical correspondences. For transformation-invariant matching, GeoTransformer~\citep{qin2022geotransformer} embeds pairwise superpoint geometry for RANSAC-free~\citep{fischler1981ransac} registration, and RoITr~\citep{yu2023rotation} addresses rotation invariance. Lepard~\citep{li2022lepard} further handles rigid/deformable partial matching. These methods are developed and evaluated on same-modality benchmarks such as 3DMatch~\citep{zeng20173dmatch}. Recent work explores geometric refinement with skip attention~\citep{zhang2026igasa}, surgical partial-to-full registration~\citep{zhou2026pointwise}, and modality transformation under limited overlap~\citep{wu2025mtpcr}. Cross-modal methods instead bridge different data forms, spanning images, point clouds, language, and CT~\citep{li2023twod3dmatr,ren2023corri2p,yuan2025seeing,10.1145/3664647.3681154}. In CT-US registration, both inputs are bone-surface point clouds, yet their sampling depends on the imaging modality. CUFormer therefore keeps geometric features shared and conditions them on CT/US identities through modality embeddings and modality-specific adapters. Sharing the geometric features allows the encoder to be pretrained on large-scale same-modality data, reducing the demand for large paired CT-US datasets.

\subsection{Adapter-based Modality Adaptation}
Parameter-efficient adaptation modifies pretrained models through small trainable modules. Representative approaches include bottleneck adapters~\citep{houlsby2019adapter} and low-rank updates~\citep{hu2022lora}. Related strategies use prompts or adapters to tune vision transformers~\citep{jia2022vpt,chen2022adaptformer} and multimodal backbones~\citep{zhou2025mettle}. These methods adapt a pretrained model to a new task or domain while keeping its backbone weights largely frozen. CUFormer instead uses adapters to condition a shared geometric matcher on input modality. It adopts the bottleneck adapter structure for modality-conditioned feature updates and combines it with modality embeddings in MAT. Separate CT and US residual branches provide modality-specific parameters while the attention layers remain shared. The matching stage trains both the shared parameters and these modules.

\section{Method}

\subsection{Overview}
\textbf{Problem setup.}
Let the preoperative CT bone surface be the \emph{reference} point cloud $\mathbf{P}=\{\mathbf{p}_i\in\mathbb{R}^3\}$ and the intraoperative US observation be the \emph{source} point cloud $\mathbf{Q}=\{\mathbf{q}_j\in\mathbb{R}^3\}$. Because bone is rigid, we seek a rigid transform $\mathbf{T}=(\mathbf{R},\mathbf{t})$, with rotation $\mathbf{R}$ and translation $\mathbf{t}$, that aligns $\mathbf{Q}$ to $\mathbf{P}$. Two properties make this harder than standard same-modality registration: (i)~$\mathbf{P}$ and $\mathbf{Q}$ come from different imaging modalities, and (ii)~$\mathbf{Q}$ covers only a small, partially visible part of the surface that $\mathbf{P}$ describes in full, so most reference points have no counterpart in the source.

\textbf{Architecture.}
As illustrated in Figure~\ref{fig:overview}, CUFormer follows a coarse-to-fine registration pipeline. A shared KPConv-FPN encoder~\citep{thomas2019kpconv} extracts multi-resolution features from the CT reference and US source. The coarse level represents superpoints and their local patches, while the fine level retains dense pointwise features. MAT (Section~\ref{sec:mat}) supplies modality identity embeddings and modality-specific residual updates within the shared geometric matcher. Reference-side visibility (Section~\ref{sec:visibility}) weights the resulting coarse scores before superpoint correspondence selection. Within matched patch pairs, visibility biases fine scores before Sinkhorn-based optimal transport establishes point correspondences. A local-to-global estimator then recovers the rigid transformation.

\subsection{Modality-Aware Transformer}
\label{sec:mat}
MAT conditions a shared geometric matcher on the input modality through identity encoding and residual adaptation. The embeddings provide stream-specific offsets to coarse and fine descriptors. The adapters provide separate CT and US residual feature transformations after the shared attention blocks.

\paragraph{Modality identity encoding.}
Shared geometric features do not explicitly encode the acquisition modality. Owing to the imaging characteristics of CT and US, the geometry of each point cloud carries modality identity information. Based on this idea, we couple modality identity with the complete-reference (CT) and partial-source (US) roles. Let $\mathbf{f}_i^c$ and $\mathbf{f}_i^f$ denote the coarse and fine features before modality encoding, and let $\tilde{\mathbf{f}}$ denote the encoded features. Two learnable embeddings, $\mathbf{e}_{\mathrm{ct}}$ and $\mathbf{e}_{\mathrm{us}}$, identify the CT reference and US source streams. At the coarse level, we add the embedding before the transformer,
\begin{equation}
    \tilde{\mathbf{f}}^{\,c}_{i} = \mathbf{f}^{\,c}_{i} + \mathbf{e}_{m(i)},
    \label{eq:embed-c}
\end{equation}
where $m(i)\in\{\mathrm{ct},\mathrm{us}\}$ identifies the modality. At the fine level, projection $\phi$ maps the same embedding to the fine-feature dimension before addition,
\begin{equation}
    \tilde{\mathbf{f}}^{\,f}_{i}=\mathbf{f}^{\,f}_{i}+\phi(\mathbf{e}_{m(i)}).
    \label{eq:embed-f}
\end{equation}
Eqs.~\ref{eq:embed-c} and~\ref{eq:embed-f} provide consistent modality cues at both correspondence levels, yielding the modality-aware features of Figure~\ref{fig:overview} (bottom left). The embeddings are initialized with small random values to avoid perturbing the initial features.

\paragraph{Modality-specific residual adaptation.}
The superpoint transformer contains $N_t=3$ self-/cross-attention pairs, comprising six attention blocks in total. Geometric Self-Attention (GeoSA) operates within each point cloud using GeoTransformer's distance-and-angle structure embedding~\citep{qin2022geotransformer}. Cross-Attention (CA) exchanges information between the CT and US point clouds. Separate CT and US bottleneck adapters follow each shared attention block, providing modality-specific feature updates beyond the embedding offsets.

Each adapter applies a gated bottleneck residual update:
\begin{equation}
    \mathrm{Adapter}(\mathbf{x}) = \mathbf{x} + \gamma\,\mathbf{W}_{\mathrm{up}}\,\mathrm{ReLU}\!\left(\mathbf{W}_{\mathrm{down}}\,\mathbf{x}\right),
    \label{eq:adapter}
\end{equation}
where $\mathbf{W}_{\mathrm{down}}$ is a down-projection layer to a low-dimensional space, ReLU denotes a Rectified Linear Unit, $\mathbf{W}_{\mathrm{up}}$ is the up-projection layer, and $\gamma$ is a scalar gate. $\gamma$ is initialized to zero~\citep{bachlechner2021rezero,zhang2023controlnet}, making each adapter an identity mapping at the beginning of training.

For each block $\ell$, the reference features pass through $\mathrm{Adapter}^{\mathrm{ct}}_{\ell}$ and the source features through $\mathrm{Adapter}^{\mathrm{us}}_{\ell}$, as shown in Figure~\ref{fig:overview} (bottom left). These layer- and modality-specific adapters, together with the embeddings of Eqs.~\ref{eq:embed-c} and~\ref{eq:embed-f}, form MAT.

\subsection{Reference-Side Visibility-Guided Matching}
\label{sec:visibility}
We model imaging visibility through geometric overlap between the CT surface and the partial source observation, using this overlap to represent which CT regions are observed. The reference-side visibility head predicts this imaging visibility, and its scores guide coarse and fine matching.

\textbf{Visibility head.}
We hierarchically fuse coarse, fine, and full-resolution reference features along the point-to-node hierarchy to obtain a feature $\mathbf{g}_i$ for each CT point. Each feature combines local geometry with global context from cross-attention. A small per-point network comprising $1\!\times\!1$ convolutions with group normalization maps this feature to a visibility logit, and a sigmoid produces a per-point visibility probability $v_i=\sigma(\mathrm{VisHead}(\mathbf{g}_i))\in[0,1]$.

\textbf{Soft visibility supervision.}
We supervise the imaging visibility prediction $v_i$ with a continuous soft target derived from geometric overlap. For each reference point, we compute $d_i=\min_j\|\mathbf{p}_i-\mathbf{T}^\star\mathbf{q}_j\|_2$ using the Ground-Truth (GT) source-to-reference transform $\mathbf{T}^\star$ and map this distance to a score
\begin{equation}
    s_i = \mathrm{clip}\!\left(1 - d_i / \rho,\; 0,\; 1\right),
    \label{eq:soft}
\end{equation}
where $\rho$ is an overlap radius. We mask points whose target lies in the uncertain band between the negative and positive thresholds. For the retained points, the targets remain continuous, preserving differences in geometric support within each supervised region. The visibility head minimizes a smooth-$L_1$ regression loss between $v_i$ and $s_i$, balancing contributions from the positive and negative groups.

\textbf{Visibility-aware matching.}
We assign full-resolution CT points to their nearest coarse or fine node and average valid visibility predictions within each group. Each aggregated probability $v$ becomes a weight
\begin{equation}
    w = \mathrm{clip}(v,\epsilon,1)^{\lambda},
    \label{eq:weight}
\end{equation}
with probability floor $\epsilon=10^{-3}$ and exponent $\lambda=1$. For $\ell_2$-normalized coarse descriptors $\mathbf{h}^{\mathrm{ct}}_i$ and $\mathbf{h}^{\mathrm{us}}_j$ of CT and US superpoints, let $A_{ij}=\exp(-\|\mathbf{h}^{\mathrm{ct}}_i-\mathbf{h}^{\mathrm{us}}_j\|_2^2)$. The coarse score $S_{ij}=(A_{ij}/\sum_k A_{ik})(A_{ij}/\sum_k A_{kj})$ is the product of the row- and column-normalized $A_{ij}$ and is therefore nonnegative. Visibility-weighted scores $w_iS_{ij}$ favor visible CT regions during global top-$K$ selection. At the fine stage, $Z$ denotes the point-pair logit matrix within a matched patch pair. Each reference point's $\lambda_f\log w$ is added to its row of point-pair scores before Sinkhorn optimal transport~\citep{cuturi2013sinkhorn,sarlin2020superglue}. Scores of the unmatched (dustbin) entries remain unchanged. The weight $\lambda_f$ controls how strongly low visibility suppresses assignments to a reference point.
\begin{figure*}[htb]
    \centering
    \includegraphics[width=\textwidth,keepaspectratio]{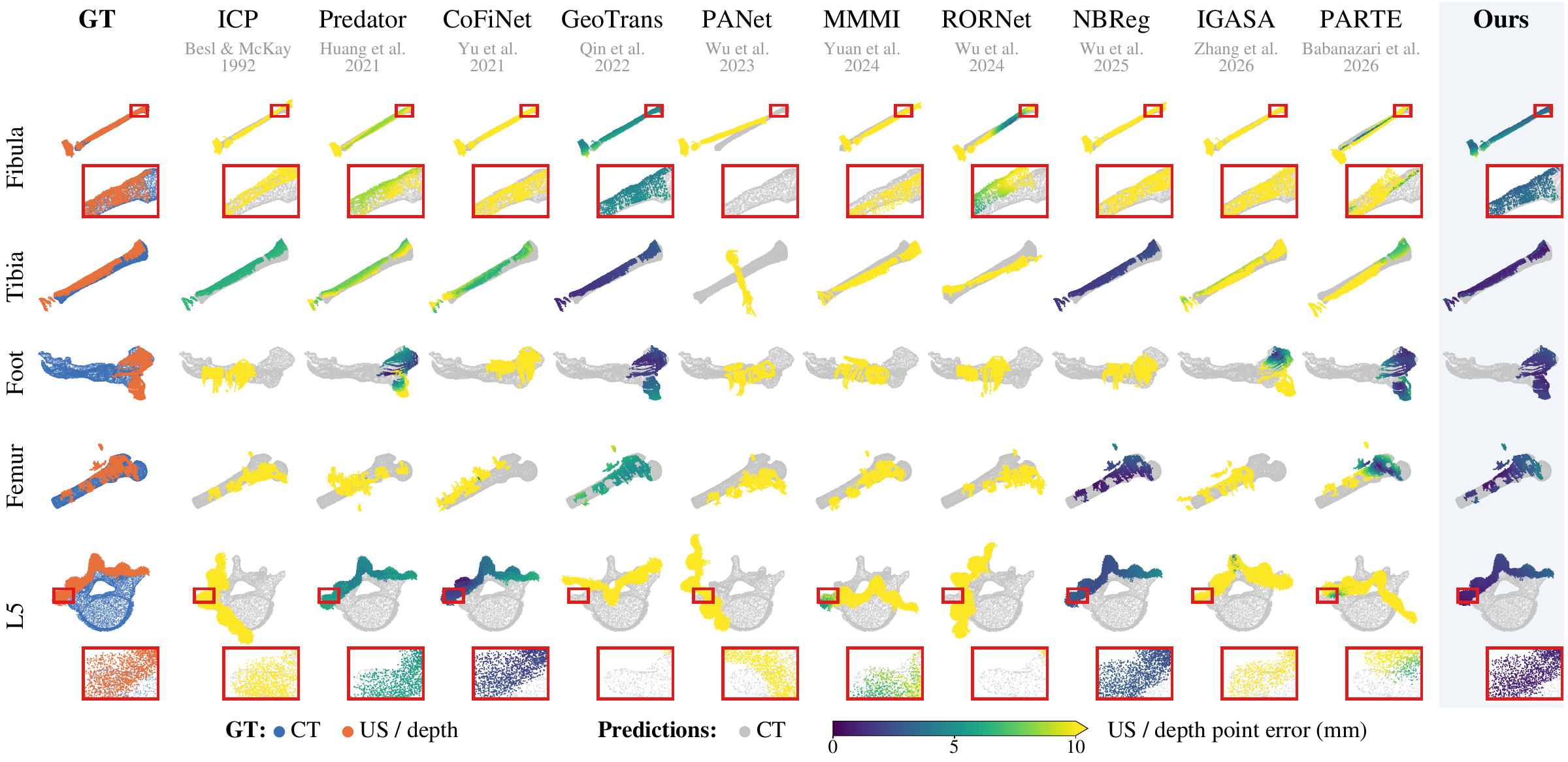}
    \caption{Qualitative visualization of registration cases. Four CT-US rows show UltraBones100k fibula, tibia, and foot and UltraBonesHip femur, followed by an additional CT-RGB-D SpineDepth L5 row. Baselines are ordered by publication year, and GeoTrans denotes GeoTransformer. GT shows CT in blue and US/depth in orange. Other columns show CT in gray and source points colored by pointwise error on a 0 to 10\,mm scale. Red-framed insets enlarge the boxed regions in the fibula and L5 rows. Ours uses MAT and visibility-guided matching.}
    \label{fig:qualitative}
\end{figure*}

\subsection{Loss and Training}
Training proceeds in two stages. Stage~1 trains the matcher, including MAT, to learn correspondences. Stage~2 freezes the encoder, shared attention, embeddings, and adapters and optimizes only the visibility head. Stage~1 uses the correspondence loss $\mathcal{L}_{\mathrm{stage1}}$ and Stage~2 uses the visibility loss $\mathcal{L}_{\mathrm{stage2}}$, defined as
\begin{equation}
    \begin{aligned}
        \mathcal{L}_{\mathrm{stage1}} &= \mathcal{L}_{\mathrm{coarse}} + \mathcal{L}_{\mathrm{fine}},\\
        \mathcal{L}_{\mathrm{stage2}} &= \lambda_{v}\,\mathcal{L}_{\mathrm{vis}}.
    \end{aligned}
    \label{eq:loss}
\end{equation}
where $\mathcal{L}_{\mathrm{coarse}}$ is a weighted circle loss on superpoint descriptors. $\mathcal{L}_{\mathrm{fine}}$ is the negative log-likelihood of GT point correspondences under the optimal-transport assignment. $\mathcal{L}_{\mathrm{vis}}$ is the smooth-$L_1$ regression loss against the soft targets in Eq.~\ref{eq:soft}, weighted by $\lambda_{v}$.

\section{Experiments}
\subsection{Experimental Setup}
\textbf{Datasets.}
We conduct experiments on two primary CT-US datasets, UltraBones100k~\citep{wu2025ultrabones100k} and UltraBonesHip~\citep{wu2025neuralbonereg}. Like US, RGB-D observes only a partial bone surface, but its acquisition differs, so we additionally evaluate on SpineDepth~\citep{liebmann2021spinedepth}. The two CT-US datasets provide CT surface models and US image sequences, covering five anatomical structures: fibula, tibia, foot, femur, and pelvis. UltraBones100k is collected from 14 specimens and contains 135 distinct scans, and UltraBonesHip is collected from five specimens and provides 62 scans. SpineDepth provides CT surface models and depth point clouds from 10 specimens, with 100 pairs.

\textbf{Preprocessing.} UltraBones100k uses released US reconstructions from GT bone labels. UltraBonesHip clouds are obtained with the automatic reconstruction script released with NBReg~\citep{wu2025neuralbonereg}. SpineDepth uses the supplied segmented depth clouds. We sample 20,000 points per CT mesh and normalize both clouds by the CT sample centroid and half its bounding-box diagonal. All three datasets provide only spatially aligned CT-US/RGB-D pairs. To simulate spatial misalignment between preoperative and intraoperative acquisitions, the main evaluation applies random rigid transformations to the US/RGB-D source clouds while keeping CT fixed. Rotations are sampled uniformly from $\mathrm{SO}(3)$, and translation components are sampled independently from $\mathcal{U}(-0.5,0.5)$ in the normalized coordinates. The GT transform maps the perturbed source back to the CT frame.

\textbf{Metrics.}
We report Relative Rotation Error (RRE, degrees), Relative Translation Error (RTE, mm), strict pointwise Root Mean Squared Error (RMSE, mm), and Flip Rate (FR). For $N$ source points $\mathbf{q}_j$, RMSE is $\sqrt{N^{-1}\sum_{j=1}^{N}\|\hat{\mathbf{T}}\mathbf{q}_j-\mathbf{T}^\star\mathbf{q}_j\|_2^2}$. Translation and point errors are converted to millimeters using half the CT bounding-box diagonal. Within each test set, RRE, RTE, and RMSE are averaged over all scans, including all failures. FR is the percentage of scans with RRE $\geq90^\circ$. FR measures large-angle failures and partially indicates anatomical flips.

\textbf{Baselines.}
We compare against two groups of baselines: per-case optimization methods and learning-based methods. The first group includes ICP~\citep{besl1992icp}, PARTE~\citep{babanazari2026parte}, and NBReg~\citep{wu2025neuralbonereg}. ICP iteratively refines a local alignment, and PARTE performs plane-assisted robust estimation from hand-crafted feature correspondences. NBReg is designed for multimodal bone-surface registration and therefore serves as our primary medical comparator. Learning-based methods include Predator~\citep{huang2021predator}, CoFiNet~\citep{yu2021cofinet}, GeoTransformer~\citep{qin2022geotransformer}, PANet~\citep{wu2023panet}, RORNet~\citep{wu2024rornet}, and IGASA~\citep{zhang2026igasa}, together with the unsupervised MMMI~\citep{yuan2024mmmi}. ICP uses the Open3D implementation, and the other baselines use their official code. We rerun the complete official NBReg pipeline, including CT-specific UDF fitting and test-time registration optimization, and evaluate it using our shared test inputs and metrics. We use three specimen-level cross-validation splits with non-overlapping test specimens. Each split holds out two specimens for testing on UltraBones100k and SpineDepth and one on UltraBonesHip, which provides only five specimens. All methods share the test samples and poses within each split, and scoring uses up to 30,000 fixed source points per scan.

\begin{figure}[htb]
    \centering
    \includegraphics[width=\columnwidth]{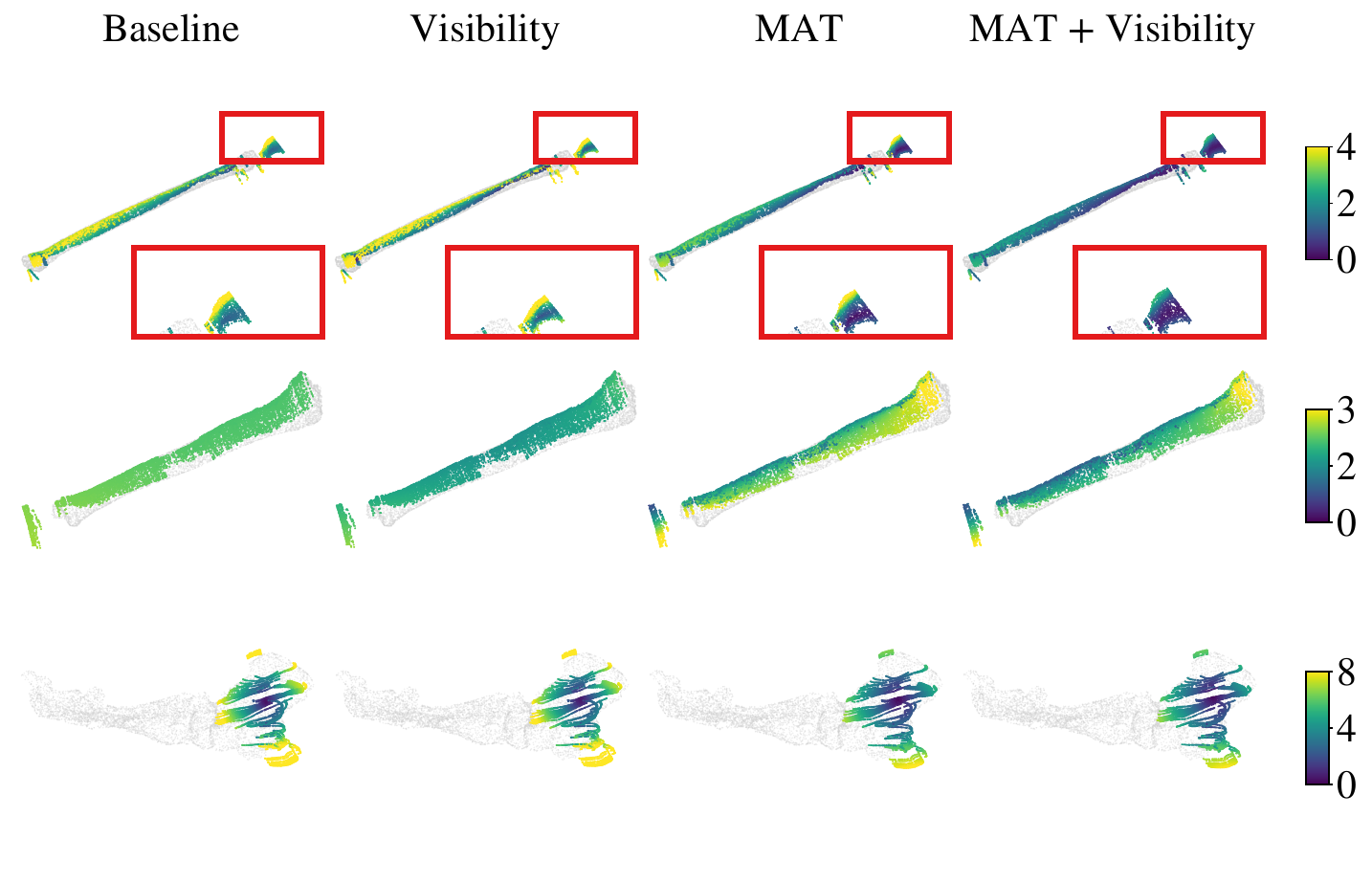}
    \caption{Qualitative comparison of the component ablation on UltraBones100k. CT is gray. US points are colored by displacement from the GT alignment in millimeters, with a shared color scale within each row. In the fibula row, insets enlarge the same red-boxed region for all methods.}
    \label{fig:ablation}
\end{figure}

\textbf{Implementation details.}
We train a separate model for each split of each dataset with the same training settings. All parameters shared with GeoTransformer~\citep{qin2022geotransformer} are initialized from its weights pretrained on 3DMatch~\citep{zeng20173dmatch}, while the modality embeddings and adapters are newly initialized. Stage~1 trains the matcher, including MAT, with Adam for 300 epochs on a single NVIDIA RTX 4090 GPU, with batch size 1 and learning rate $5\times10^{-5}$. Each adapter uses a bottleneck reduction ratio $r=4$. Stage~2 trains the visibility head for 15 epochs with $\lambda_v=1$, overlap radius $\rho=0.05$ in normalized coordinates, and positive/negative thresholds of 0.6/0.2. Visibility is fused at both matching stages with fine-stage weight $\lambda_f=0.5$. On SpineDepth, the RGB-D surfaces are processed by the source branch with its modality embedding and adapters.

\begin{table*}[t]
\centering
\caption{CT-US comparison with additional CT-RGB-D evaluation on SpineDepth. Results are mean $\pm$ sample standard deviation across the three specimen-level cross-validation splits. Each split averages over its test scans, including failures. RRE is in degrees, RTE and RMSE are in mm, and FR is in \%. Lower is better, and best means are bold. GeoTrans denotes GeoTransformer. The upper group contains per-case optimization methods, and the lower group contains learning-based methods, each ordered by publication year.}
\label{tab:main}
\footnotesize
\newcommand{\tablestd}[1]{\kern0.3pt{\normalfont\fontsize{5}{5}\selectfont$\pm$\kern0.3pt#1}}
\setlength{\tabcolsep}{0.15pt}
\renewcommand{\arraystretch}{1.04}
\begin{tabular*}{\textwidth}{@{\extracolsep{\fill}}l*{12}{r}@{}}
\toprule
Method & \multicolumn{4}{c}{UltraBones100k: CT-US} & \multicolumn{4}{c}{UltraBonesHip: CT-US} & \multicolumn{4}{c}{SpineDepth: CT-RGB-D} \\
\cmidrule(lr){2-5}\cmidrule(lr){6-9}\cmidrule(lr){10-13}
 & RRE $\downarrow$ & RTE $\downarrow$ & RMSE $\downarrow$ & FR $\downarrow$ & RRE $\downarrow$ & RTE $\downarrow$ & RMSE $\downarrow$ & FR $\downarrow$ & RRE $\downarrow$ & RTE $\downarrow$ & RMSE $\downarrow$ & FR $\downarrow$ \\
\midrule
ICP & 84.16\tablestd{7.02} & 64.46\tablestd{4.62} & 112.58\tablestd{14.71} & 50.8\tablestd{2.75} & 97.86\tablestd{8.19} & 57.34\tablestd{1.52} & 85.36\tablestd{13.54} & 53.0\tablestd{6.94} & 91.67\tablestd{11.61} & 20.02\tablestd{1.69} & 36.50\tablestd{0.64} & 55.0\tablestd{10.00} \\
NBReg & 41.25\tablestd{9.30} & 28.87\tablestd{3.49} & 35.21\tablestd{2.03} & 27.0\tablestd{2.75} & 29.11\tablestd{13.74} & 18.84\tablestd{6.70} & 19.45\tablestd{4.97} & 19.7\tablestd{9.46} & 10.97\tablestd{5.51} & 2.97\tablestd{1.32} & 4.21\tablestd{1.70} & 6.7\tablestd{2.89} \\
PARTE & 66.70\tablestd{9.70} & 25.21\tablestd{4.63} & 40.02\tablestd{7.84} & 34.9\tablestd{7.27} & 92.35\tablestd{23.36} & 47.53\tablestd{9.03} & 54.80\tablestd{10.87} & 53.0\tablestd{18.37} & 128.25\tablestd{4.79} & 32.75\tablestd{2.74} & 42.84\tablestd{0.84} & 81.7\tablestd{5.77} \\
\midrule
Predator & 16.09\tablestd{0.45} & 7.14\tablestd{0.40} & 7.42\tablestd{0.10} & 1.6\tablestd{2.75} & 58.07\tablestd{20.02} & 27.30\tablestd{2.54} & 32.22\tablestd{7.03} & 27.3\tablestd{12.03} & 5.65\tablestd{0.67} & 3.57\tablestd{0.22} & 3.48\tablestd{0.14} & \textbf{0.0}\tablestd{0.00} \\
CoFiNet & 38.99\tablestd{10.50} & 16.74\tablestd{4.64} & 30.64\tablestd{20.28} & 15.9\tablestd{7.27} & 77.00\tablestd{13.91} & 35.61\tablestd{9.87} & 57.50\tablestd{9.19} & 37.9\tablestd{6.94} & 5.06\tablestd{0.24} & 2.98\tablestd{0.33} & 3.16\tablestd{0.22} & \textbf{0.0}\tablestd{0.00} \\
GeoTrans & 19.00\tablestd{4.32} & 13.19\tablestd{0.14} & 13.56\tablestd{0.39} & 1.6\tablestd{2.75} & 26.04\tablestd{9.90} & 16.31\tablestd{2.00} & 19.09\tablestd{4.31} & 6.1\tablestd{6.94} & 21.58\tablestd{4.37} & 5.34\tablestd{1.27} & 7.82\tablestd{1.61} & 11.7\tablestd{2.89} \\
PANet & 94.84\tablestd{14.52} & 78.72\tablestd{6.26} & 143.23\tablestd{18.64} & 55.6\tablestd{7.27} & 89.30\tablestd{3.95} & 63.69\tablestd{3.41} & 90.30\tablestd{11.67} & 47.0\tablestd{6.94} & 87.05\tablestd{13.43} & 35.45\tablestd{2.03} & 49.97\tablestd{2.24} & 50.0\tablestd{5.00} \\
MMMI & 85.47\tablestd{7.10} & 51.74\tablestd{0.10} & 87.37\tablestd{0.98} & 41.3\tablestd{5.50} & 96.94\tablestd{15.21} & 45.44\tablestd{4.42} & 65.31\tablestd{2.30} & 50.0\tablestd{13.64} & 118.63\tablestd{1.62} & 22.50\tablestd{0.60} & 43.24\tablestd{1.54} & 85.0\tablestd{8.66} \\
RORNet & 88.19\tablestd{10.91} & 46.92\tablestd{2.45} & 87.20\tablestd{18.06} & 46.0\tablestd{7.27} & 89.90\tablestd{2.82} & 59.64\tablestd{4.46} & 75.21\tablestd{5.75} & 48.5\tablestd{5.25} & 101.33\tablestd{19.02} & 28.59\tablestd{3.45} & 38.67\tablestd{5.00} & 53.3\tablestd{20.82} \\
IGASA & 51.66\tablestd{3.48} & 15.28\tablestd{1.43} & 20.36\tablestd{5.54} & 22.2\tablestd{2.75} & 98.58\tablestd{4.91} & 49.80\tablestd{1.95} & 46.94\tablestd{1.28} & 57.6\tablestd{6.94} & 61.46\tablestd{16.15} & 10.49\tablestd{1.52} & 21.87\tablestd{4.04} & 31.7\tablestd{7.64} \\
\midrule
\textbf{CUFormer} & \textbf{4.53}\tablestd{0.83} & \textbf{3.19}\tablestd{0.33} & \textbf{2.54}\tablestd{0.10} & \textbf{0.0}\tablestd{0.00} & \textbf{22.00}\tablestd{2.92} & \textbf{12.55}\tablestd{1.48} & \textbf{15.07}\tablestd{1.41} & \textbf{1.5}\tablestd{2.62} & \textbf{4.72}\tablestd{5.02} & \textbf{2.18}\tablestd{0.93} & \textbf{2.36}\tablestd{1.28} & \textbf{0.0}\tablestd{0.00} \\
\bottomrule
\end{tabular*}
\end{table*}

\subsection{Comparison with State-of-the-Art}
\label{sec:additional-datasets}

Table~\ref{tab:main} reports the CT-US comparisons and the additional CT-RGB-D results on SpineDepth under the same protocol.

\textbf{CT-US registration.} CUFormer achieves the lowest mean errors and FR on both CT-US datasets, with 4.53$^\circ$, 3.19\,mm, 2.54\,mm, and FR 0.0\% on UltraBones100k and 22.00$^\circ$, 12.55\,mm, 15.07\,mm, and FR 1.5\% on UltraBonesHip. The baselines show distinct failure patterns across method groups. ICP and PARTE exceed 66$^\circ$ mean RRE on both datasets, with FR of 34.9--53.0\%. This is consistent with ICP's need for a close initialization and with the unreliable hand-crafted correspondences available to PARTE under partial US coverage. NBReg ends in large-angle failures on some cases, with FR of 27.0\% and 19.7\%. Differences from its reported performance may reflect initial poses, test partitions, and anatomical coverage, as our UltraBones100k evaluation also includes foot scans, which its original evaluation excludes. On UltraBonesHip, its RMSE of 19.45\,mm is comparable to the 19.09\,mm of GeoTransformer. Among learning-based methods, Predator and GeoTransformer are the strongest baselines on UltraBones100k, with RMSE of 7.42 and 13.56\,mm and FR of 1.6\%. On UltraBonesHip, where the hip anatomy is deeper and harder to image with US~\citep{wu2025neuralbonereg}, errors increase for the three learning-based baselines with the lowest RMSE on UltraBones100k. Predator's RMSE rises from 7.42 to 32.22\,mm and its FR from 1.6\% to 27.3\%, and IGASA's RMSE more than doubles. GeoTransformer shows the smallest increase among them, from 13.56 to 19.09\,mm. CUFormer also degrades but retains the lowest RMSE. Under controlled initial misalignment on one UltraBones100k split, CUFormer also has the lowest mean RMSE at every perturbation level, as detailed in Supplementary Section~S2.

\textbf{Additional CT-RGB-D evaluation.} On SpineDepth, CUFormer obtains mean RRE 4.72$^\circ$, RTE 2.18\,mm, RMSE 2.36\,mm, and FR 0.0\%. Unlike on the CT-US datasets, several baselines also perform well: CoFiNet and Predator reach RMSE of 3.16 and 3.48\,mm with FR 0.0\%, and NBReg reaches 4.21\,mm. The margin narrows markedly, possibly because spinal geometry has less end-for-end ambiguity than structures such as the tibia. CUFormer still reduces RMSE by 25.3\% relative to CoFiNet.

Figure~\ref{fig:qualitative} shows five qualitative cases, and CUFormer keeps low errors in all of them. On the fibula, ICP, CoFiNet, NBReg, and IGASA place the US points on the CT surface but reach the 10\,mm limit of the error scale, consistent with misalignment along the near-symmetric shaft rather than gross misplacement. PANet places the tibia scan across the bone axis. On the foot, ICP, CoFiNet, PANet, MMMI, RORNet, and NBReg place the US points on a different part of the foot from the GT. GeoTransformer and NBReg align the tibia well, NBReg is close to CUFormer on the femur, and CoFiNet and NBReg are accurate on L5, where Predator has moderate errors.

\begin{figure}[htb]
    \centering
    \includegraphics[width=\columnwidth]{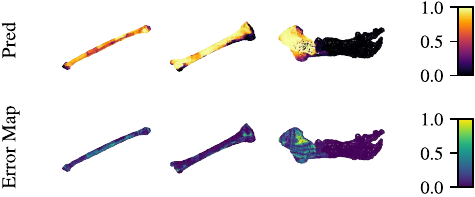}
    \caption{Predicted reference-side visibility. Top: predicted probability $v_i$. Bottom: deviation $|v_i - s_i|$ from the soft target. The prediction separates the US-observed part of the CT surface from the unobserved remainder.}
    \label{fig:visibility}
\end{figure}

\subsection{Ablation Study}
The ablations in Tables~\ref{tab:ablation-modal} and~\ref{tab:ablation-vis} use the same three splits and training settings as the main comparison, with each configuration trained and evaluated per split and averaged across splits. Table~\ref{tab:ablation-modal} compares Baseline, MAT, MAT without pretraining, Visibility, and MAT + Visibility on UltraBones100k. Baseline uses the shared geometric matcher with modality embeddings, adapters, and visibility disabled. MAT adds the embeddings and adapters, while Visibility adds visibility guidance with MAT disabled. MAT without pretraining uses the same configuration as MAT, but its shared parameters are not initialized from the 3DMatch weights. Each module improves Baseline on its own: MAT reduces RMSE from 13.56 to 5.41\,mm, and Visibility reduces it to 7.88\,mm. MAT outperforms Visibility on all three metrics, showing that distinguishing the two modalities within the shared matcher yields the larger single-module gain. Without pretraining, RMSE rises from 5.41 to 16.60\,mm, about 3.1$\times$ that of MAT, and RTE and RMSE are the highest in Table~\ref{tab:ablation-modal}. This indicates that initializing the shared parameters from same-modality pretraining is critical when paired CT-US data are limited. In relative terms, the benefit of visibility guidance does not diminish once MAT is added: on top of MAT, it reduces RMSE by 53.0\%, from 5.41 to 2.54\,mm, compared with 41.9\% on Baseline, and further lowers RRE by 2.53$^\circ$ and RTE by 2.71\,mm. This is consistent with the two modules addressing different challenges, namely modality-dependent sampling and asymmetric coverage. Combining both modules gives the best results. The qualitative examples in Figure~\ref{fig:ablation} follow the same trend. On the fibula, Visibility largely retains the shaft errors of Baseline, whereas MAT lowers them and the combination gives the lowest errors. On the tibia, MAT leaves higher errors toward one end of the scan, which visibility guidance reduces. On the foot, the differences are smaller, and the combination shows the fewest high-error points.

\begin{table}[htb]
    \centering
    \caption{Component ablation on UltraBones100k, averaged across three splits. w/o pretrain: shared parameters are not initialized from 3DMatch weights. RRE is in degrees, and RTE and RMSE are in mm.}
    \label{tab:ablation-modal}
    \footnotesize
    \setlength{\tabcolsep}{4pt}
    \begin{tabular}{l ccc}
        \hline
        Setting & RRE $\downarrow$ & RTE $\downarrow$ & RMSE $\downarrow$ \\
        \hline
        Baseline & 19.00 & 13.19 & 13.56 \\
        + MAT & 7.06 & 5.90 & 5.41 \\
        + MAT (w/o pretrain) & 15.10 & 15.46 & 16.60 \\
        + Visibility & 7.95 & 7.54 & 7.88 \\
        \textbf{+ MAT + Visibility} & \textbf{4.53} & \textbf{3.19} & \textbf{2.54} \\
        \hline
    \end{tabular}
\end{table}

\subsection{Ablation of Visibility-Guided Matching}
Table~\ref{tab:ablation-vis} evaluates the incremental benefit of visibility fusion over the MAT configuration, labeled No Visibility. Coarse weighting prioritizes CT patches, and fine weighting adjusts point assignments within them, including competition with the dustbin. Either stage alone yields most of the gain: coarse-only fusion reduces RMSE from 5.41 to 2.85\,mm, and fine-only fusion to 2.79\,mm. Fusing visibility at both stages gives the lowest errors among the predicted-visibility configurations. The remaining gap to the GT-visibility oracle is small, at 0.02$^\circ$, 0.34\,mm, and 0.19\,mm, so fusion at both stages recovers 99.2\%, 88.9\%, and 93.8\% of the oracle's improvement over No Visibility in RRE, RTE, and RMSE. These results support fusion at both stages.

To assess the prior used for matching, we pool reference points from the UltraBones100k test scans of one randomly selected split and label them with the training thresholds, excluding the uncertain band. Average Precision (AP) is 92.04\%, Area Under the Receiver Operating Characteristic Curve (AUROC) is 94.81\%, and F1 is 86.33\% at $v_i\geq0.5$. Expected Calibration Error (ECE) is 0.025 over 15 equal-width bins on $[0,1]$, weighting each bin's absolute gap between mean probability and positive-label frequency by its fraction of labeled points. The high AP and low ECE indicate that the predictions both separate visible from unobserved regions and are well calibrated, consistent with the small gap to the GT-visibility oracle in Table~\ref{tab:ablation-vis}. In Figure~\ref{fig:visibility}, deviations from the target are small overall, and larger deviations appear mainly near the boundary between observed and unobserved regions, as on the foot, where intermediate targets fall in the uncertain band excluded from supervision.

\begin{table}[htb]
    \centering
    \caption{Ablation of visibility-guided matching on top of MAT, averaged across three splits. Coarse and Fine fuse predicted visibility at the respective matching stage. Full fuses it at both stages. GT uses GT visibility as an oracle. RRE is in degrees, and RTE and RMSE are in mm.}
    \label{tab:ablation-vis}
    \footnotesize
    \begin{tabular}{l ccc}
        \hline
        Setting & RRE $\downarrow$ & RTE $\downarrow$ & RMSE $\downarrow$ \\
        \hline
        No Visibility & 7.06 & 5.90 & 5.41 \\
        + Visibility Coarse & 4.85 & 3.65 & 2.85 \\
        + Visibility Fine & 4.81 & 3.60 & 2.79 \\
        \textbf{+ Visibility Full} & \textbf{4.53} & \textbf{3.19} & \textbf{2.54} \\
        \hline
        + Visibility GT & 4.51 & 2.85 & 2.35 \\
        \hline
    \end{tabular}
\end{table}

\section{Conclusion}
We presented CUFormer, a modality-aware CT-US bone-surface registration framework for orthopedic navigation. It combines a Modality-Aware Transformer and reference-side visibility-guided matching in a shared geometric matcher to address two key challenges: (i) modality-dependent surface sampling, which hinders CT-US matching, and (ii) asymmetric coverage from acoustic shadowing, which makes correspondences ambiguous on near-symmetric bones. Sharing the geometric encoder lets CUFormer inherit pretraining on large-scale same-modality data, which the ablations show to be critical, and as a learned matcher it registers new cases without per-case optimization. CUFormer achieves the lowest mean errors and FR on UltraBones100k and UltraBonesHip, and results on SpineDepth support its use for CT-RGB-D registration. These results suggest that modeling modality differences and imaging visibility on shared geometric features is a promising basis for reliable US-based orthopedic navigation. Future work will advance the method toward intraoperative clinical use.

\textbf{Limitations.} UltraBonesHip retains larger residual errors than the other evaluated datasets, with mean RMSE 15.07\,mm. The method assumes rigid anatomy and requires GT transforms for visibility supervision. Generalization to unseen anatomy and acquisition settings, and prospective intraoperative validation, remain open.

\clearpage
\section*{LLM Usage Statement}
The research idea and CUFormer implementation were developed by the authors. OpenAI GPT-6-Astra assisted with manuscript preparation, including language polishing and figure preparation, literature searches, and running and reproducing comparison baselines. OpenAI GPT-Image-2 was used to create the illustrative ultrasound probe icon in Figure~\ref{fig:teaser}. The authors are responsible for the scientific claims, experimental results, and references in this paper.

{\small\bibliographystyle{ieeenat_fullname}\bibliography{main}}

\clearpage
\setcounter{section}{0}\setcounter{figure}{0}\setcounter{table}{0}
\renewcommand{\thesection}{S\arabic{section}}\renewcommand{\thetable}{S\arabic{table}}\renewcommand{\thefigure}{S\arabic{figure}}
\renewcommand{\theHsection}{supp.\arabic{section}}\renewcommand{\theHtable}{supp.\arabic{table}}\renewcommand{\theHfigure}{supp.\arabic{figure}}
\maketitlesupplementary
\section{Computational Efficiency}
Table~\ref{tab:supp-efficiency} summarizes measured registration times for CUFormer and NeuralBoneReg (NBReg) on NVIDIA RTX 4090 GPUs. The CUFormer measurements use the checkpoints and inference settings of main-paper Table~1, with five repetitions per test scan. They cover the model forward pass, including feature extraction, correspondence matching, and pose estimation within the model.

\begin{table}[t]
\centering
\caption{Measured times in seconds on the held-out test sets of one randomly selected split per dataset, which for UltraBones100k is the split used for the initial-pose evaluation. CUFormer reports model-forward timings. NBReg reports complete intraoperative optimization measured with shared GPUs. UDF fitting, model loading, and warmup are outside the timing intervals.}
\label{tab:supp-efficiency}
\small
\setlength{\tabcolsep}{3pt}
\begin{tabular}{lrrrr}
\toprule
& \multicolumn{2}{c}{CUFormer} & \multicolumn{2}{c}{NBReg} \\
Dataset & Mean & Median & Mean & Median \\
\midrule
UltraBones100k & 0.0552 & 0.0500 & 16.04 & 16.05 \\
UltraBonesHip & 0.1089 & 0.1007 & 23.44 & 23.47 \\
SpineDepth & 0.1151 & 0.1014 & 13.87 & 14.00 \\
\bottomrule
\end{tabular}
\end{table}

NBReg reuses a fitted and preloaded CT Unsigned Distance Field (UDF), with one complete registration per test scan. UltraBones100k, UltraBonesHip, and SpineDepth contain 21, 22, and 20 scans, respectively. Timing includes source centering, tensor construction and GPU transfer, pose-network and optimizer initialization, global candidate scoring, candidate refinement, 1,000 optimization steps, and final transformation output. UltraBones100k and SpineDepth use 2,000 candidates, while UltraBonesHip uses 5,000. Each step scores all candidates on 512 points and refines the scores of the best 64 using up to 3,000 points. UDF fitting and loading are completed before timing, consistent with preoperative preparation.

Four RTX 4090 GPUs each run one NBReg worker, processing assigned scans sequentially while sharing the devices with existing workloads. Each worker performs one complete unmeasured warmup registration per dataset. CUDA synchronization brackets each timing interval. Each registration uses one GPU, with the four workers processing different scans. The across-scan sample standard deviations for NBReg are 1.54, 0.10, and 0.24 seconds on the three datasets, respectively. These timings characterize the stated configurations and timing scopes, with their absolute values affected by shared GPU load. Measurements with matched preprocessing scope and exclusive devices would further quantify the runtime difference.

\section{Initial-Pose Robustness on UltraBones100k}
\begin{figure*}[t]
    \centering
    \includegraphics[width=\textwidth]{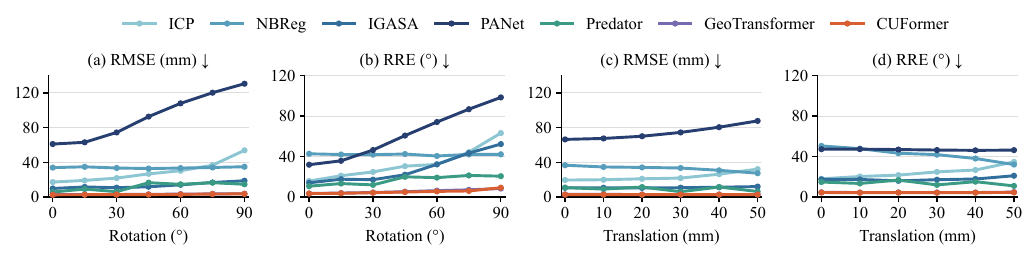}
    \caption{Robustness of seven methods to initial misalignment on the UltraBones100k test scans of one randomly selected split. The US is rotated about the CT centroid and translated while the CT stays fixed. (a,b) Mean strict RMSE and mean RRE versus rotation at 30\,mm translation. (c,d) The same metrics versus translation at 30$^\circ$ rotation.}
    \label{fig:supp-initial-pose}
\end{figure*}

Figure~\ref{fig:supp-initial-pose} compares CUFormer, ICP, NBReg, IGASA, PANet, Predator, and GeoTransformer on the UltraBones100k test scans of one randomly selected split, the same split as the main-paper visibility evaluation. In this separate controlled experiment, CT remains fixed as in the main evaluation, and only US is perturbed. Starting from the aligned CT-US pairs, we form the perturbed source $\mathbf{Q}'=\mathbf{E}\mathbf{Q}$ while retaining the CT reference $\mathbf{P}$, so the source-to-reference ground truth is $\mathbf{E}^{-1}$. Rotation is about the fixed CT sample centroid, and translation is the displacement applied to US in physical millimeters. The rotation sweep uses 0$^\circ$ to 90$^\circ$ in 15$^\circ$ steps at 30\,mm translation, and the translation sweep uses 0 to 50\,mm in 10\,mm steps at 30$^\circ$ rotation. Each sweep therefore varies one component under a fixed nonzero offset in the other. Five rotation-axis/translation-direction pairs per scan are held fixed across perturbation levels and shared across methods. The 30$^\circ$/30\,mm condition is reused in both sweeps, giving $(7+6-1)\times5=60$ unique perturbations per scan and method.

Each perturbation is evaluated with the retained main-comparison checkpoint of this split or the method's inference recipe, without retraining. Method-specific inference settings and NBReg's 1,000-step optimization are retained. Three additional identity-input checks per method are excluded from the curves. All metrics use the original fixed main-table evaluation source pool, with distances converted by half the CT bounding-box diagonal. RRE measures orientation error, and RMSE measures displacement of the sampled bone surface. FR counts RRE $\geq90^\circ$, and SR requires both RRE $<5^\circ$ and RTE $<4$\,mm. Repeated directions are not independent specimens.

CUFormer has the lowest mean RMSE at every perturbation level in both sweeps and the lowest mean RRE except at 90$^\circ$, where GeoTransformer and CUFormer reach 8.47$^\circ$ and 9.29$^\circ$, respectively. The methods respond to the initial pose in three distinct ways. The errors of ICP, IGASA, PANet, and Predator grow with rotation: from 0$^\circ$ to 90$^\circ$, RMSE rises from 17.33 to 53.90\,mm for ICP, from 9.93 to 18.84\,mm for IGASA, from 61.13 to 130.64\,mm for PANet, and non-monotonically from 5.84 to 14.78\,mm for Predator. The errors of ICP and PANet also grow with translation. NBReg keeps RMSE between 32.91 and 34.96\,mm across all rotations, close to its three-split mean RMSE of 35.21\,mm under uniform $\mathrm{SO}(3)$ rotations in main-paper Table~1. CUFormer and GeoTransformer change little with translation and degrade only gradually with rotation. From 0$^\circ$ to 90$^\circ$, the RMSE of CUFormer rises from 2.64 to 3.70\,mm, and its RRE increases mainly at the upper end, from 6.05$^\circ$ at 75$^\circ$ to 9.29$^\circ$ at 90$^\circ$. Within this range, GeoTransformer and CUFormer reach RMSE of 3.89 and 3.70\,mm at 90$^\circ$, a much smaller gap than in main-paper Table~1, where their three-split mean RMSE is 13.56 and 2.54\,mm. Because main-paper Table~1 samples rotations uniformly from $\mathrm{SO}(3)$, about 81.8\% of its rotation angles exceed 90$^\circ$, beyond the range of this sweep. This suggests that the errors of GeoTransformer in main-paper Table~1 arise mainly from larger rotations, whereas CUFormer keeps the lowest mean RMSE in both settings.

At 90$^\circ$/30\,mm, CUFormer has SR 53.3\% and FR 1.0\%. At 30$^\circ$/50\,mm, its mean RMSE is 2.85\,mm, SR is 54.3\%, and FR is 0.0\%. At the largest rotation, ICP, NBReg, IGASA, PANet, and Predator reach FR of 35.2\%, 25.7\%, 17.1\%, 60.0\%, and 4.8\%, respectively. This FR of 1.0\% is not directly comparable with the 0.0\% FR of CUFormer in main-paper Table~1, which uses a different perturbation protocol and all three splits. FR detects large-angle errors only and must be interpreted alongside continuous errors and success statistics. These robustness results are specific to the evaluated perturbation protocol and specimen sample.

\section{Architecture Figure Details}
Main-paper Figure~3 shows the shared encoder, MAT, and reference-side visibility-guided matching, with CT in blue and US in orange. Point clouds, visibility maps, feature colors from joint Principal Component Analysis (PCA) of MAT features, and coarse score blocks are real intermediate outputs on an UltraBones100k tibia test case. In the correspondence panel, US is at the estimated pose and offset for display. Green lines are the strongest visibility-weighted coarse matches and gray lines show fine matches. Bottom left: modality embeddings condition the coarse descriptors before attention and the fine descriptors before matching. Modality-specific gated adapters follow each shared geometric self-attention and cross-attention block. Bottom right: the visibility head fuses coarse features with KPConv fine and full-resolution features into CT visibility $v_i$. Brighter regions indicate greater predicted geometric support from US. Soft targets $s_i$ derive from the distance $d_i$ to the GT-aligned US. The weights $w$ scale coarse scores $S$ before top-$K$ selection, and $\lambda_f\log w$ biases fine scores $Z$ before Sinkhorn. Fine-score cells are schematic. Selected patch pairs share MAT-feature colors. Dashed arrows indicate training signals. Stage 2 freezes the matcher and trains only the visibility head using visibility supervision. Fine-stage visibility fusion uses stopped gradients.
\end{document}